\documentclass[sigconf]{acmart}
\usepackage{array}
\usepackage{tabularx}
\usepackage{multirow}
\AtBeginDocument{%
  }

\setcopyright{acmlicensed}
\copyrightyear{2025}
\acmYear{2025}
\acmDOI{XXXXXXX.XXXXXXX}
\acmConference[ACM SIGSPATIAL 2025]{International Conference on Advances in Geographic Information Systems}{November 2025}{Minneapolis, MN, USA}
\acmISBN{978-1-4503-XXXX-X/2018/06}

\renewcommand\footnotetextcopyrightpermission[1]{}
\begin{document}

\title[A Unified Framework for Next-Gen Urban Forecasting]{A Unified Framework for Next-Gen Urban Forecasting via LLM-driven Dependency Retrieval and GeoTransformer}

\author{Yuhao Jia}
\email{yuhao.jia@emory.edu}
\affiliation{
  \institution{Emory University, University of Pennsylvania}
  \country{USA}
}

\author{Zile Wu}
\email{wuzile@alumni.upenn.edu}
\affiliation{
  \institution{University of Pennsylvania}
  \city{Philadelphia}
  \country{USA}
}

\author{Shengao Yi}
\email{shengao@upenn.edu}
\affiliation{
  \institution{University of Pennsylvania}
  \city{Philadelphia}
  \country{USA}
}

\author{Yifei Sun}
\email{sophiasun@alumni.upenn.edu}
\affiliation{
  \institution{University of Pennsylvania}
  \city{Philadelphia}
  \country{USA}
}

\author{Xiao Huang}
\email{xiao.huang2@emory.edu}
\affiliation{
  \institution{Emory University}
  \city{Atlanta}
  \country{USA}
}
\renewcommand{\shortauthors}{Jia et al.}

\begin{abstract}
Urban forecasting has increasingly benefited from high-dimensional spatial data through two primary approaches: graph-based methods which rely on predefined spatial structures, and region-based methods that focus on learning expressive urban representations. Although these methods have laid a strong foundation, they either rely heavily on structured spatial data, struggle to adapt to task-specific dependencies, or fail to integrate holistic urban context. Moreover, no existing framework systematically integrates these two paradigms and overcome their respective limitations.
To address this gap, we propose a novel, unified framework for high-dimensional urban forecasting, composed of three key components: (1) the Urban Region Representation Module that organizes latent embeddings and semantic descriptions for each region, (2) the Task-aware Dependency Retrieval module that selects relevant context regions based on natural language prompts, and (3) the Prediction Module, exemplified by our proposed GeoTransformer architecture, which adopts a novel geospatial attention mechanism to incorporate spatial proximity and information entropy as priors.
Our framework is modular and supports diverse representation methods and forecasting models, and can operate even with minimal input. Quantitative experiments and qualitative analysis across six urban forecasting tasks demonstrate strong task generalization and validate the framework’s effectiveness.
\end{abstract}




\maketitle

\section{Introduction}

In urban forecasting tasks, classical methods usually rely statistical and machine learning methods that operate on low-dimensional, hand-engineered features \cite{li2022predicting,roskam2008predictive, dhs2013demographic,10422038,moreira2013predicting,pappalardo2013understanding}.  While effective in constrained settings, these approaches struggle to model the complexity of urban systems.

Recent advances in spatial representation learning, remote sensing, and deep neural architectures have introduced a new paradigm in urban modeling: transforming urban regions into high-dimensional latent representations to better capture complex urban dynamics. Such representations are commonly derived from text embedding \cite{chen2022_points_of_interest_relationship_inference_spatial_enriched, li2020competitive}, spatial representation learning \cite{mai2023spatial,wu2024torchspatial, li2023urban} or by encoding satellite imagery data \cite{jean2018Tile2Vec, wang2024deep}.

High-dimensional urban forecasting applications can be broadly categorized into two directions. The first utilizes graph-based modeling with spatial feature embeddings, then using Graph Neural Networks (GNNs) or Graph Attention Networks (GATs) for predictions \cite{feng2022adaptive, zheng2020gman, chen2022_points_of_interest_relationship_inference_spatial_enriched, li2020competitive, li2023urban}. While effective, these methods depend heavily on predefined spatial structures and high-quality spatial data,  which limits their flexibility in data-sparse or dynamically changing environments.
The second direction focuses on region-based methods, which derive high-dimensional representations directly from satellite imagery or other high-resolution spatial data \cite{jean2018Tile2Vec, wang2024deep, satmaepp2024rethinking}. These methods produce compact representations that preserve built environment features and support downstream tasks. However, these approaches only utilize local information within each patch for prediction and lack the capability to incorporate global urban context \cite{wang2024deep}, which is crucial for tasks requiring holistic understanding. 

The limitations and incompatibility of the two paradigms ultimately reflect a structural divergence rooted in whether spatial dependency is available—either built into the input or entirely absent.
Several studies have explored automated mechanisms for capturing spatial dependencies for high-dimensional representations, including spatial autocorrelation, proximity, or sparse regression \cite{fu2019efficient,regionEncoder, li2020forecaster}. However, these approaches remain task-agnostic.
To date, no unified framework exists that systematically integrates the two modeling paradigms through task-aware dependency modeling to address their respective limitations.

To address these gaps, we propose a novel, unified and modular framework for high-dimensional urban forecasting. It consists of three functional modules: (1) the Urban Region Representation Module encodes each region into high-dimensional embeddings and semantic descriptions; (2) the Task-Aware Dependency Retrieval Module identifies spatial dependencies among regions by matching task-specific prompts with semantic descriptions; and (3) the Prediction Module aggregates embeddings of retrieved regions for final prediction.

Each module is technically decoupled but logically aligned, enabling flexible integration and replacement of existing models. It supports multimodal encoding methods, but also remains effective without predefined spatial structures, requiring no more than satellite imagery in minimal settings. The framework automatically captures task-specific spatial dependencies through a language-driven retrieval process, making it theoretically applicable to any static urban forecasting task.

While the framework supports diverse prediction models, most existing decoding methods rarely consider how characteristics of high-dimensional urban representations affect information aggregation. To address this, we introduce GeoTransformer, a transformer-based architecture equipped with a novel geospatial attention mechanism, which incorporates spatial proximity and information entropy as priors to weight cross-attention.

We validate the effectiveness of our framework through extensive quantitative experiments and qualitative analysis across six urban forecasting tasks.

Our main contributions can be summarized as follows:
\begin{itemize}
    \item We propose a unified, modular framework for high dimensional urban forecasting, composed of urban region representation, task-aware dependency retrieval, and a prediction module.
    \item We introduce GeoTransformer, a transformer-based prediction module that integrates spatial proximity and information entropy as priors to guide cross-attention.
    \item We demonstrate the framework’s flexibility and effectiveness across six urban forecasting tasks through quantitative and qualitative evaluation.
\end{itemize}

\section{Related Work}

\begin{figure*}[h]
\centering
\includegraphics[width=1\textwidth]{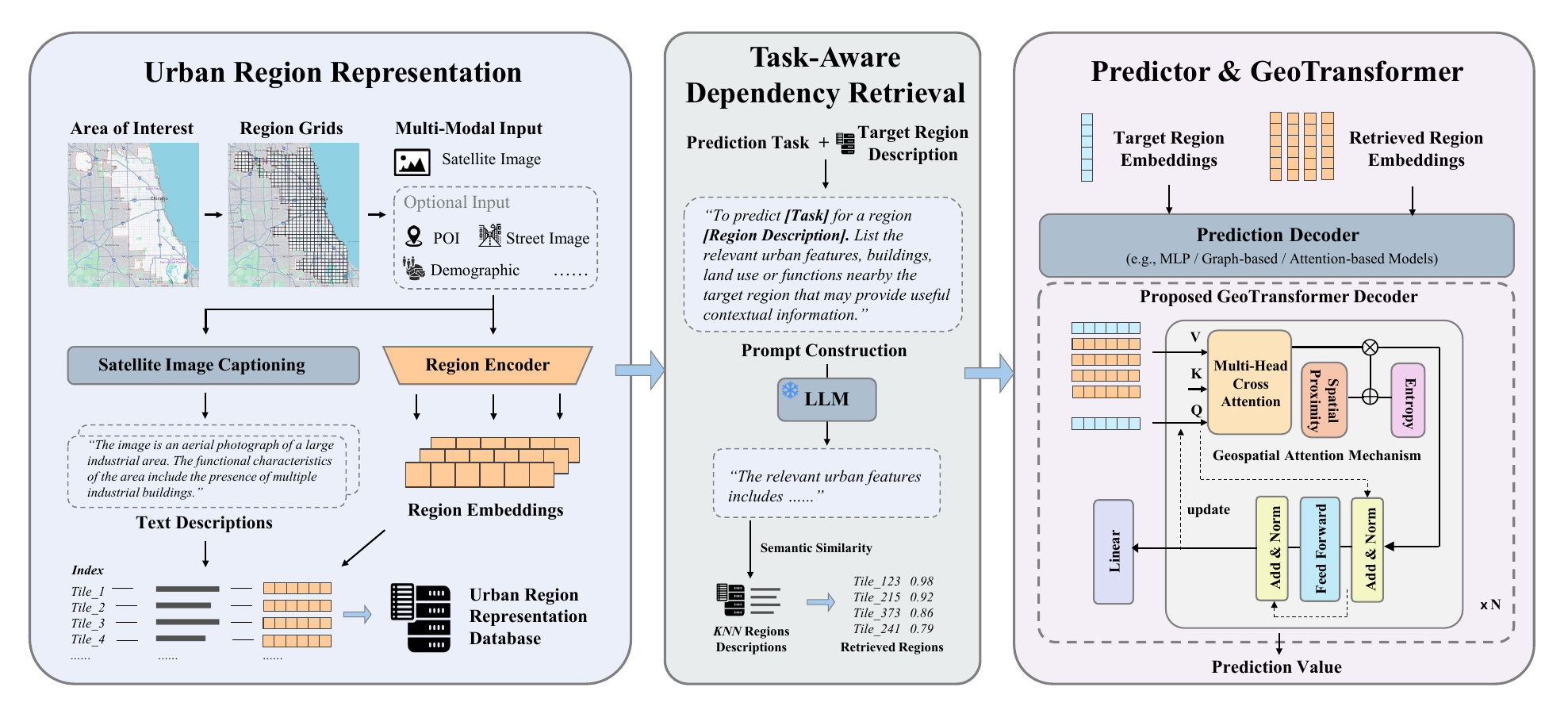}
\caption{Our high-dimensional urban forecasting framework composed of urban region representation, task-aware dependency retrieval and prediction module.}
\label{fig:framework}
\end{figure*}

\subsection{High-Dimensional Representations for Urban Forecasting}
Urban forecasting has increasingly leveraged high-dimensional data to address complex challenges. Traditional approaches often rely on low-dimensional numerical data, such as Point of Interest (POI) data\cite{li2022predicting}, survey data\cite{roskam2008predictive, dhs2013demographic}, GPS records \cite{10422038,moreira2013predicting,pappalardo2013understanding}, demographic census, spatial features and so forth, limiting their ability to capture the multifaceted nature of urban dynamics. Recent advances have focused on two primary directions for high-dimensional urban representations.

The first direction employs graph-based methods that embed urban information into representations by constructing predefined graph structures. Graph Attention Networks (GATs) \cite{Graph_Attention_Networks} are used to infer POI relationships \cite{chen2022_points_of_interest_relationship_inference_spatial_enriched, li2020competitive, li2023urban}.
\citeauthor{feng2022adaptive} propose an Adaptive Graph Spatial-Temporal Transformer Network to model cross-spatial-temporal correlations. \citeauthor{zheng2020gman} use a graph multi-attention network to model the impact of the spatio-temporal factors on traffic conditions. \citeauthor{chen2022_points_of_interest_relationship_inference_spatial_enriched} propose a spatial-aware attention module based on spatial proximity. Additionally, position embeddings are employed as learnable parameters to provide location information \cite{xu2021spatialtemporaltransformernetworkstraffic}. These methods rely heavily on predefined spatial structures, such as road networks or adjacency matrices, which limits their applicability in dynamic or data-sparse urban environments. 

The second direction focuses on region-based methods that leverage satellite imagery and other high-resolution data to encode urban areas into high-dimensional representations. Tile2Vec \cite{jean2018Tile2Vec} learns representations from satellite imagery tiles. \citeauthor{satmaepp2024rethinking} applies masked autoencoding to learn representations in a self-supervised manner \cite{satmaepp2024rethinking}. \citeauthor{wang2024deep} propose a deep hybrid model fusing regional built environment and socio-demographic information into latent representations through multi-task learning \cite{wang2024deep}. These methods transform imagery into compact latent representations that capture rich spatial characteristics of the built environment, and have been shown effective in urban prediction tasks. However, these methods only leverage local information patches for prediction and struggle to incorporate global urban context or dynamically model spatial dependencies. 

\subsection{Automatic Spatial Dependency Modeling}
The reliance of graph-based methods on predefined spatial structures, the locality constraint of region-based representations, and the incompatibility between the two paradigms can all be traced to a fundamental difference in the availability of spatial dependency information. 
Several studies have proposed automatic mechanisms to capture spatial dependencies for high-dimensional spatial embeddings. \citeauthor{fu2019efficient} leverage top-K locality and spatial autocorrelation  to capture influence weights across regions \cite{fu2019efficient}. RegionEncoder connects regions based on spatial proximity \cite{regionEncoder}. \citeauthor{li2020forecaster} apply sparse regression to construct spatial dependency \cite{li2020forecaster}. However, all of these methods remain task-agnostic, providing the same dependency structure regardless of the prediction objective—despite the potential misalignment with task-specific spatial relevance. For example, sparse regression captures linear dependencies in feature space, assigning higher weights to regions with similar characteristics. In the context of predicting ride-share demand for a residential area, such a method may prioritize other residential zones over nearby transit infrastructure, failing to reflect domain-relevant influence.

Recent advances in large language models \cite{Bommasani2021FoundationModels} and remote sensing image captioning techniques \cite{cheng2024domain, huang2024understanding} offer a new possibility for reasoning about spatial dependencies at the semantic level, enabling adaptation to diverse prediction tasks. However, this direction remains unexplored.

\subsection{Characteristics and Aggregation of High-Dimensional Region Representations}
Existing methods rarely examine how the properties of high dimensional urban representations affect information aggregation and propagation. Specifically, some region-based methods generally treat all representations equally, without considering the variation in information richness or predictive utility across regions\cite{wang2024deep}. 
Additionally, some graph-based methods incorporate geospatial priors (such as spatial distance) as weights or bias into self-attention mechanisms. However, they often overlook a key limitation of self-attention: representations in later layers have already undergone substantial aggregation and may no longer correspond to their original spatial locations \cite{chen2022_points_of_interest_relationship_inference_spatial_enriched, li2023urban}. Applying priors to such misaligned tokens is spatially inconsistent and theoretically unsound.

Few studies have examined these issues or investigated how architecture design should be adapted to the structural and semantic properties of high-dimensional urban region representations.

\section{Methodology}
\subsection{Framework Overview}
As illustrated in Figure \ref{fig:framework}, we present a unified and modular framework for high-dimensional urban forecasting, composed of three functionally decoupled modules:  (1) the Urban Region Representation Module, which encodes each city region into high-dimensional embeddings and semantic descriptions; (2) the Task-Aware Dependency Retrieval Module, which selects relevant context regions based on LLM-inferred task-specific prompts; and (3) the Prediction Module, which aggregates the target and retrieved regions information for final prediction.

The first and last module are designed to be interchangeable, allowing flexible integration of different representation encoders or prediction architectures. The framework does not require predefined spatial structures and can operate with minimal inputs (satellite imagery). 

Building on this structure, we further propose GeoTransformer, a transformer-based decoder designed to address limitations in existing forecasting models. It integrates spatial proximity and information entropy into attention computation to enhance the training efficiency and prediction performance.

\subsection{Urban Region Representation}
The Urban Region Representation Module encodes each region into two forms of representation: a high-dimensional latent embedding $z_i$ for numerical prediction and a semantic description $d_i$ for task-aware retrieval. Satellite imagery is required as the primary input, ensuring spatial consistency and enabling visual encoding across regions. Additional urban data can be incorporated depending on the chosen encoding method. Formally, we define the encoding process as:
\begin{equation}
    z_i = f(I_i,u_i) \label{e1}
\end{equation}
where $I_i$ is the satellite image of region $i$, $u_i$ denotes optional auxiliary urban features, and $f(\cdot)$ represents a flexible encoder that maps the input to a latent space.

To enable language-based reasoning in dependency retrieval, a semantic description $d_i$ is also generated from satellite imagery using a remote sensing captioning model:
\begin{equation}
    d_i = \text{Desc}(I_i)
\end{equation}
where $\text{Desc}(\cdot)$ produces a natural language summary of the region's built environment.

Together, $z_i$ and $d_i$ are stored as a centralized region representation database, providing a standardized interface for retrieval and prediction modules.

\subsection{Task-aware Dependency Retrieval}
As illustrated in Figure~\ref{fig:framework}, we propose a language-driven retrieval module to identify task-relevant spatial dependencies.

For each forecasting task and a target region $i$, we construct a natural language prompt using the task description and the region’s description $d_i$:
\begin{quote}
To predict [Task] for a given target region described as follows: [Region Description]. 

List the relevant urban features, buildings, land use or functions nearby the target region that may provide useful contextual information.
\end{quote}
This prompt is then passed to a pre-trained large language model (LLM), which infers a textual description $r_i$ representing the prototype of regions likely to influence the target region's prediction.

To incorporate spatial locality, we use $k$-nearest neighbors (k-NN) as a spatial constraint to define the candidate region set, following common practice in urban spatial modeling \cite{fu2019efficient}. Specifically, for each region $i$, $\text{k-NN}(i)$ denotes the $k$ regions that are spatially closest to the target region $i$ based on Euclidean distance. The value of $k$ is a tunable hyperparameter, often informed by the spatial resolution of the study area, such as the typical size of neighborhoods or planning units.

For each region $j$ within the candidate set $\text{k-NN}(i)$, we compute the semantic similarity between the expected context region prototype $r_i$ and the region’s semantic description $d_j$. We then select the top-$n$ most similar regions as the final context set:
\begin{equation}
Z_i = \{ z_j \mid j \in \operatorname{argTop}_n(\text{Sim}(r_i, d_j)) \cap \text{k-NN}(i) \}
\end{equation}
where set $Z_i$ represents the $n$ retrieved context regions for the target region $i$, with $n$ treated as a tunable hyperparameter.

The similarity function $\text{Sim}(\cdot, \cdot)$ defines semantic matching between textual descriptions and can accommodate various implementations; In this work, we use sentence-level embeddings (encoded by BGE-M3\cite{bge-m3}) and cosine similarity for efficiency and consistency. 

As LLMs encode a broad base of human knowledge and spatial understanding \cite{Bommasani2021FoundationModels}, this retrieval process can be interpreted as an automated proxy for human inference when identifying relevant contextual spatial dependencies. Moreover, because the mechanism operates purely at the semantic level, it is theoretically generalizable across a wide range of urban forecasting tasks.

\renewcommand{\arraystretch}{1.2}
\begin{table*}[t]
\centering
\newcolumntype{L}[1]{>{\hsize=#1\hsize}X}
\begin{tabularx}{0.9\textwidth}{L{0.35}|L{0.65}}
\hline
\multicolumn{2}{l}{\rule[-1.5ex]{0pt}{4ex}\textbf{\textit{Panel 1 Region-based Encoding Methods}}} \\ \hline
Model 1: Tile2Vec & representation dim = 512, fully connect MLP, layer=1 \\ \hline
Model 2: SatMAE\textsuperscript{++} & representation dim = 1024, fully connect MLP, layer=1 \\ \hline
Model 3: DHM & representation dim = 4096, Fully Connect MLP, layer=2  \\ \hline
\multicolumn{2}{l}{\rule[-1.5ex]{0pt}{4ex}\textbf{\textit{Panel 2 Traditional Dependency Modeling + Graph-based Methods}}} \\ \hline
Model 4: SatMAE\textsuperscript{++} + GAT (grid) & representation dim = 4096, heads = 16, attention layers = 4-5, KNN=121, others task-tuned \\ \hline
Model 5: SatMAE\textsuperscript{++} + GAT (sparse) & representation dim = 4096, heads = 16, attention layers = 4-5, KNN=121, subset size = 81, Lasso regularization $\lambda$ = 0.01, others task-tuned \\ \hline
\multicolumn{2}{l}{\rule[-1.5ex]{0pt}{4ex}\textbf{\textit{Panel 3 Our Framework}}} \\ \hline
Model 6: SatMAE\textsuperscript{++} + GAT & representation dim = 4096, heads = 16, attention layers = 4-5, KNN=121, retrieval size = 81, others task-tuned\\ \hline
Model 7: SatMAE\textsuperscript{++} + GeoTransformer & representation dim = 4096, heads = 16, transformer layers = 4-5, KNN=121, retrieval size = 81, others task-tuned \\ \hline
\end{tabularx}
\caption{Configurations of baselines and variants of our framework.} 
\label{tab:model_config} 
\end{table*}

\subsection{Prediction Module and the GeoTransformer Architecture}
The prediction module serves as the final stage of our framework, responsible for aggregating information from the target region and its retrieved context regions to generate task-specific forecasts.

Given the target region's embedding $z_i$ and the set of retrieved region embeddings $Z_i$, the module produces the final prediction $\hat{y}_i$ via a decoder function $\mathcal{D}$:
\begin{equation}
\hat{y}_i = \mathcal{D}(z_i, Z_i)
\end{equation}

This component is modular and supports a wide range of decoder architectures. For instance, fully connected networks or transformer-based decoders can process the retrieved region embeddings directly, while graph-based decoders may treat each region as a node and construct a task-specific local graph based on the retrieved dependencies. The prediction module thus acts as a flexible interface that bridges upstream dependency retrieval with downstream forecasting tasks.

Although existing methods have demonstrated strong performance in various tasks, they often overlook structural limitations inherent to high-dimensional urban representations. As discussed in Section 2, these include: (1) unequal informativeness across of region embeddings, and (2) spatial distortion caused by applying spatial priors uniformly across attention layers. 

To address these challenges, we propose GeoTransformer, a novel transformer-based decoder designed specifically for high-dimensional region representations. It introduces a geospatial attention mechanism that uses cross-attention to capture cross-region context while allowing spatial priors to be consistently applied across layers. Formally, as in the general decoder design, it maps the target region region embedding $z_i$ and its retrieved region embeddings $Z_i$ to a prediction output:
\begin{equation}
\hat{y}_i = \mathcal{G}(z_i, Z_i)
\end{equation}

As shown in the prediction module in Figure \ref{fig:framework}, the first layer of the model applies a geospatial attention mechanism that computes cross-attention between the target and retrieved representations, with the attention scores weighted by spatial proximity and information entropy. The attention calculation is represented as:
\begin{equation}
    \text{GeoAtt}(Q,K,V) = \{ \alpha W_S + (1-\alpha) W_E\} \odot \text{softmax}(\frac{QK^T}{\sqrt{d_k}}) V
\end{equation}
where the query matrix $Q$ contains the representation of the target region $z_i$, while the value matrix $V$ consists of $z_i$ itself and the retrieved region representations as value regions. The key matrix $K$ is set as trainable weights. $d_k$ denotes the dimension of the key vectors. 
$W_s$ and $W_e$ are the spatial proximity and information entropy weighting factors respectively. $\alpha \in [0, 1]$ is a balancing coefficient that controls the relative importance of spatial proximity versus information entropy in attention weighting.

The design of geospatial attention mechanism leverages cross-attention not only to model the interaction between the target region and its retrieved related regions, but more importantly, to preserve spatial alignment across layers. Unlike self-attention, where multi-layer propagation fuses all token information and breaks spatial correspondence, our design naturally updates the query token in each layer while keeping the value tokens fixed to the original region embeddings. This allows spatial priors to be consistently applied across all layers.
In addition, we parameterize the key matrix $K$ as trainable weights rather than tying it to the value representations. This design allows the evolving query representations to be matched against keys in the same latent space, avoiding mismatch between abstract queries and fixed, low-level value embeddings.
Finally, by computing attention only for a single query token, the cross-attention design also reduces memory and computational cost compared to full self-attention over all regions.

\textbf{Spatial proximity} weighting is adopted to provide location information. Based on the assumption of Tobler's First Law of Geography 
\cite{tobler1970computer}, we assigned regions closer to the target region with higher weights. We leverage a linear weighting method, scaling the distances within 0 to 1, inversely transforming shorter distances into higher weights:
  \begin{equation}
    W_{S_j} = 1 - \frac{d_j}{\text{max}(d)}
\end{equation}
where $d_j$ is the distance to the $j^{th}$ value region, and $\text{max}(d)$ are the maximum distances observed.

\textbf{Information entropy} has been shown to evaluate the effectiveness of high-dimensional data \cite{wu2024feature} and has been used for weighting \cite{zhu2020effectiveness}. We assign higher weights for representations with higher information entropy. Regions with higher entropy are believed to contain richer and more complex urban information, thereby playing a more significant role in prediction. The calculation can be presented as:
\begin{equation}
    W_{E_j} =  \frac{H_j}{\max(H_1, H_2, \dots, H_n)}
\end{equation}
where $H_i$ is the Shannon entropy of the latent representation $z_i$, and $\max(H_1, H_2, \dots, H_n)$ is the maximum entropy across retrieved tiles. Since the the latent representation is a high-dimensional vector $z_i$, the entropy $H_i$ for each individual latent representation can be represented as:
\begin{equation}
    {H}_j = - \sum_{i=1}^{d} p_{ji} \log(p_{ji})
\end{equation}
\begin{equation}
    \textbf{where } \quad \text{p}_{ij} = \frac{e^{z_{ij}}}{\sum_{k=1}^{d} e^{z_{ik}}}
\end{equation}
Here, $z_{ji}$ is the value of the $j^{th}$ feature for latent representation $z_i$, and $p_{ij}$ is the normalized probability.

Building on these foundations, GeoTransformer adopts the standard multi-head attention structure \cite{c:22} to jointly capture diverse spatial relevance patterns:
\begin{equation}
    \text{MultiHead}(Q,K,V) = \text{Concat}(head_1, ..., head_h)W^O
\end{equation}
\begin{equation}
    \textbf{where } \text{head}_i = \text{GeoAtt}(QW^Q_i,KW^K_i,VW^V_i)
\end{equation}
where $W^Q_i$, $W^K_i$, $W^V_i$ and $W^O$ are the projection matrices. The amount of heads and layers of the decoder module is also adaptive. At last, the outputs of the last layer are passed through a fully connected linear layer to generate the final prediction. 


\renewcommand{\arraystretch}{1.2}
\begin{table*}[t]
\centering
\newcolumntype{L}{>{\raggedright\arraybackslash}m{0.22\textwidth}} 
\newcolumntype{Y}{>{\centering\arraybackslash}X}
\begin{tabularx}{1\textwidth}{LYYYYYY} 
\toprule
\multirow{2}{*}{Model} & \multicolumn{6}{c}{$R^2$} \\ 
\cmidrule(lr){2-7}
 & GDP & Housing Price & Ride-share & Traffic Crashes & Crimes & Services \\ 
\midrule
\multicolumn{7}{@{}l@{}}{\textit{Panel 1 Region-based Encoding Methods }} \\
\midrule
1 Tile2Vec & 0.484/0.320 & 0.504/0.341 & 0.551/0.498 & 0.427/0.318 & 0.427/0.284 & 0.675/0.645 \\ 
2 SatMAE\textsuperscript{++} & 0.616/0.403 & 0.757/0.558 & 0.719/0.550 & 0.689/0.425 & 0.608/0.473 & 0.836/0.769 \\
3 DHM & 0.721/0.493 & 0.923/0.326 & 0.668/0.571 & 0.813/0.281 & 0.712/0.121 & 0.856/0.212 \\
\midrule
\multicolumn{7}{@{}l@{}}{\textit{Panel 2 Traditional Dependency Modeling + Graph-based Methods}} \\
\midrule
4 SatMAE\textsuperscript{++}+ GAT (grid) & 0.700/0.435 & 0.882/0.570 & 0.817/0.501 &  0.539/0.465 & 0.594/0.521 & 0.787/0.745 \\ 
5 SatMAE\textsuperscript{++}+ GAT (sparse) &  0.754/0.561 & 0.815/0.431 & 0.781/0.694 &  0.672/0.325 & 0.419/0.211 & 0.891/0.726 \\
\midrule
\multicolumn{7}{@{}l@{}}{\textit{Panel 3 Our Framework}} \\
\midrule
6 SatMAE\textsuperscript{++}+ GAT & 0.801/0.612 & 0.787/0.641 & 0.825/0.771 & 0.773/0.453 & 0.619/0.520 & 0.832/0.797 \\
7 SatMAE\textsuperscript{++}+ GeoTransformer & \textbf{0.811/0.783} & \textbf{0.923/0.912} & \textbf{0.920/0.901} & \textbf{0.716/0.638} & \textbf{0.669/0.597} & \textbf{0.891/0.824} \\ 
\bottomrule
\end{tabularx}
\caption{Predictive performance of baselines and our framework. Each entry is represented as training/testing performance.} 
\label{tab:eval_results} 
\end{table*}

\section{Experiment}
In this section, we evaluate our framework and compare it with baseline methods on six downstream urban prediction tasks, including GDP, housing price, ride-share demand, traffic crashes, crimes and municipal service demand. To demonstrate the framework’s effectiveness, we further test its compatibility with different encoder and decoder modules. Ablation experiments are conducted to demonstrate the effectiveness of the dependency retrieval module and two weighting methods in the geospatial attention module.

\subsection{Experiment Setup}

\subsubsection{Experimental Design.}
\mbox{}\\
To evaluate the effectiveness of our proposed framework, we design a progressive comparison across three panels. The first two panels serve as baselines: Panel 1 evaluates region encoding methods, while Panel 2 builds upon the best-performing encoder and integrates task-agnostic dependency construction with graph-based decoders. The final panel applies our complete framework. By controlling the input representation across all models, this design enables a fair and systematic assessment of how our framework improves forecasting performance over existing alternatives.

Panel 1 includes three foundational urban region representation methods—Tile2Vec \cite{jean2018Tile2Vec}, SatMAE\textsuperscript{++}\cite{satmaepp2024rethinking}, and the Deep Hybrid Model (DHM) \cite{wang2024deep}. The resulting representations are directly passed to a fully connected multilayer perceptron (MLP) for prediction. This panel serves as a baseline to assess the performance of region-based methods.

Panel 2 controls for the region encoding by using a fixed representation (we adopt SatMAE\textsuperscript{++} based on its stable performance in Panel 1) and introduces spatial dependency modeling through two task-agnostic approaches. Both approaches first identify the k-nearest spatial neighbors for each target region. The first connects rook-adjacent neighbors within the candidate set to form a locally structured grid graph. The second builds upon this grid and applies sparse regression to select a subset of informative regions, which are then connected directly to the target region.
Both graphs are processed using a Graph Attention Network (GAT) decoder \cite{Graph_Attention_Networks}. This panel evaluates the combination of traditional automatic dependency modeling and graph-based models under fixed representations.

Panel 3 implements our full framework under the same fixed region representation used in Panel 2. We evaluate two variants: the first combines region-based representations with graph-based aggregation by applying a GAT decoder, while the second uses our proposed GeoTransformer decoder to realize the full version of the framework with the best overall performance. This final panel demonstrates how our framework improves forecasting outcomes.

In addition to the main comparison panels, we further evaluate the modular compatibility of our framework. To assess encoder flexibility, we use GeoTransformer as the fixed decoder and vary the region encoder across Tile2Vec, SatMAE\textsuperscript{++}, and DHM. To test decoder flexibility, we use SatMAE\textsuperscript{++} as the fixed encoder and compare GAT and GeoTransformer as decoding modules. These experiments are conducted on three representative tasks—GDP, ride-share demand, and crimes—to validate the framework’s plug-and-play capability across diverse urban prediction scenarios.

\subsubsection{Evaluation Metrics.}
\mbox{}\\
We employ commonly used statistical metrics that evaluate the accuracy of the predictions, including Mean Squared Error (MSE), Mean Absolute Error (MAE), and R-squared ($R^2$). MSE evaluates the average squared differences between predicted and actual values, MAE measures the average absolute differences, and ($R^2$) assesses the proportion of variance explained by the model. While all three metrics are computed, we report $R^2$ in the main results to ensure consistency and comparability across tasks with different value ranges.

\subsubsection{Data Preparation.}
\mbox{}\\
For the satellite imagery data, we utilized the National Agriculture Imagery Program (NAIP) four-band remote sensing imagery for the Greater Chicago Area, acquired in September 2019. The imagery has a high spatial resolution of 0.6 meters. Using Google Earth Engine, we download the dataset and subsequently divide it into tiles of size 512x512 to facilitate our analysis.

Among our six urban forecasting tasks, the GDP variable is derived from Global 1km×1km gridded revised data \cite{Chen2022}  which is based on night-time light data from DMSP/OLS and NPP/VIIRS sensors; the remaining five tasks—housing price, ride-share demand, traffic crashes, crimes and municipal service demand—are constructed from official datasets provided by the City of Chicago data portal. All datasets are selected from the same year with satellite imagery to ensure temporal consistency.

To spatially align structured labels with satellite imagery, we generate a 500m×500m fishnet grid over the study area. Spatial join and interpolation techniques are used to assign region-level values to each tile, ensuring consistent geospatial resolution across all data sources.

\subsubsection{Model Training.} 
\mbox{}\\
For the region encoders, Tile2Vec and SatMAE\textsuperscript{++} are publicly available models and are used directly without modification. The Deep Hybrid Model (DHM) is implemented and trained following the procedures described in the original paper. 
For the image captioning component in the retrieval module, we use a Qwen2-based vision-language model fine-tuned on remote sensing imagery \cite{cheng2024domain}.
All decoders—including both GAT-based models in Panel 2 and our proposed GeoTransformer—are trained using mean squared error (MSE) as the loss function across all tasks. 
Qwen2-7B-Instruct is leveraged as the LLM to infer task-relevance prototype \cite{yang2024qwen2technicalreport}.
Model-specific configurations, such as embedding dimensions, number of layers, attention heads, neighborhood size and so forth, are summarized in Table~\ref{tab:model_config}. 
All tasks are evaluated under an 80\%/20\% train-test split.
All models are trained using an NVIDIA RTX 4090 GPU running Ubuntu 22.04.

\renewcommand{\arraystretch}{1.2}
\begin{table}[t]
\centering
\begin{tabular}{lccc}
\toprule
\multirow{2}{*}{Model Variant} & \multicolumn{3}{c}{R$^2$} \\
\cmidrule(lr){2-4}
 & GDP & Ride-share & Crimes \\
\midrule
\multicolumn{4}{@{}l@{}}{\textit{Encoder Compatibility}} \\
Tile2Vec + GeoTransformer & 0.65/0.55 & 0.72/0.69 & 0.60/0.53 \\
SatMAE\textsuperscript{++}+ GeoTransformer & 0.81/0.78 & 0.92/0.90 & 0.67/0.59  \\
DHM + GeoTransformer & 0.87/0.83 & 0.91/0.89 & 0.85/0.34\\
\midrule
\multicolumn{4}{@{}l@{}}{\textit{Decoder Compatibility}} \\
SatMAE\textsuperscript{++}+ GAT & 0.80/0.61 & 0.83/0.77 & 0.62/0.52 \\
SatMAE\textsuperscript{++}+ GeoTransformer & 0.81/0.78 & 0.92/0.90  & 0.67/0.60 \\
\bottomrule
\end{tabular}
\caption{Framework compatibility experiments on three representative tasks. Each entry shows training/testing performance.}
\label{tab:compatibility_results}
\end{table}

\subsection{Evaluation Results}
\subsubsection{Main Experiment}
\mbox{}\\
Table~\ref{tab:eval_results} summarizes the model performance across six urban forecasting tasks. Among region-based baselines in Panel 1, SatMAE\textsuperscript{++} consistently outperforms other encoders, providing stable representations across all tasks. Building upon this, introducing spatial dependency modeling via GAT decoders (Panel 2) yields noticeable improvements in most tasks, but they exhibit poor compatibility across tasks due to their reliance on fixed, task-agnostic structures.

Further gains are observed when incorporating our framework in Panel 3. With the same SatMAE\textsuperscript{++} encoder, Model 7 outperforms both purely region-based and graph-based with fixed dependency methods. For instance, test $R^2$ in housing price prediction rises from 0.558 (Model 2) to 0.912, and from 0.570 (Model 4) to 0.912. Similar gains are found in ride-share demand and GDP, demonstrating the advantage of spatial context inferred from our task-aware retrieval mechanism over predefined graphs or latent similarity.

Within our framework, replacing the GAT decoder with GeoTransformer (Model 6 → Model 7) further improves prediction across all tasks. For example, in ride-share demand prediction, the test $R^2$ increases from 0.771 to 0.901. This gain reflects the effectiveness of GeoTransformer in modulating context aggregation, addressing the limitations of uniform graph aggregation. Notably, GeoTransformer exhibits strong generalization and stable performance across diverse domains, demonstrating both architectural robustness and task compatibility.

\subsubsection{Modular Compatibility}
\mbox{}\\
We assess the modularity of our framework by substituting different encoders and decoders, while keeping the other components fixed. Table~\ref{tab:compatibility_results} shows that the framework remains effective across all configurations, demonstrating a high degree of plug-and-play compatibility.

On the encoder side, all three region representations methods can be seamlessly integrated with the same decoder. Despite differences in dimensionality and input structure, each encoder enables effective downstream prediction. For example, DHM achieves the highest test accuracy in GDP prediction but overfits in crime prediction, while SatMAE\textsuperscript{++} maintains stable performance across all tasks. This relationship is also evident in Table \ref{tab:eval_results} Panel 1, where the MLP decoder applied to these three encoders yields markedly different prediction accuracy. These patterns indicate that while our framework is structurally compatible with diverse encoders, overall performance still reflects the stability and informativeness of the underlying representations.

On the decoder side, we compare GeoTransformer with GAT under a fixed encoder. Both decoders yield valid and reasonably strong predictions across all tasks, confirming that the prediction module is architecturally decoupled from upstream components. While GeoTransformer generally achieves higher accuracy, both models function effectively within the framework, underscoring its flexibility.

These results confirm that our design supports independent replacement of modules—encoders may differ in modality or dimensionality, and decoders can be upgraded or simplified as needed, all without altering or retraining the rest of the pipeline. This modular compatibility enables the framework to accommodate new representation methods and forecasting architectures while preserving overall stability.

\subsection{Qualitative Analysis}
\subsubsection{Task-aware Retrieval}
\begin{figure}[h]
\centering
\includegraphics[width=0.45\textwidth]{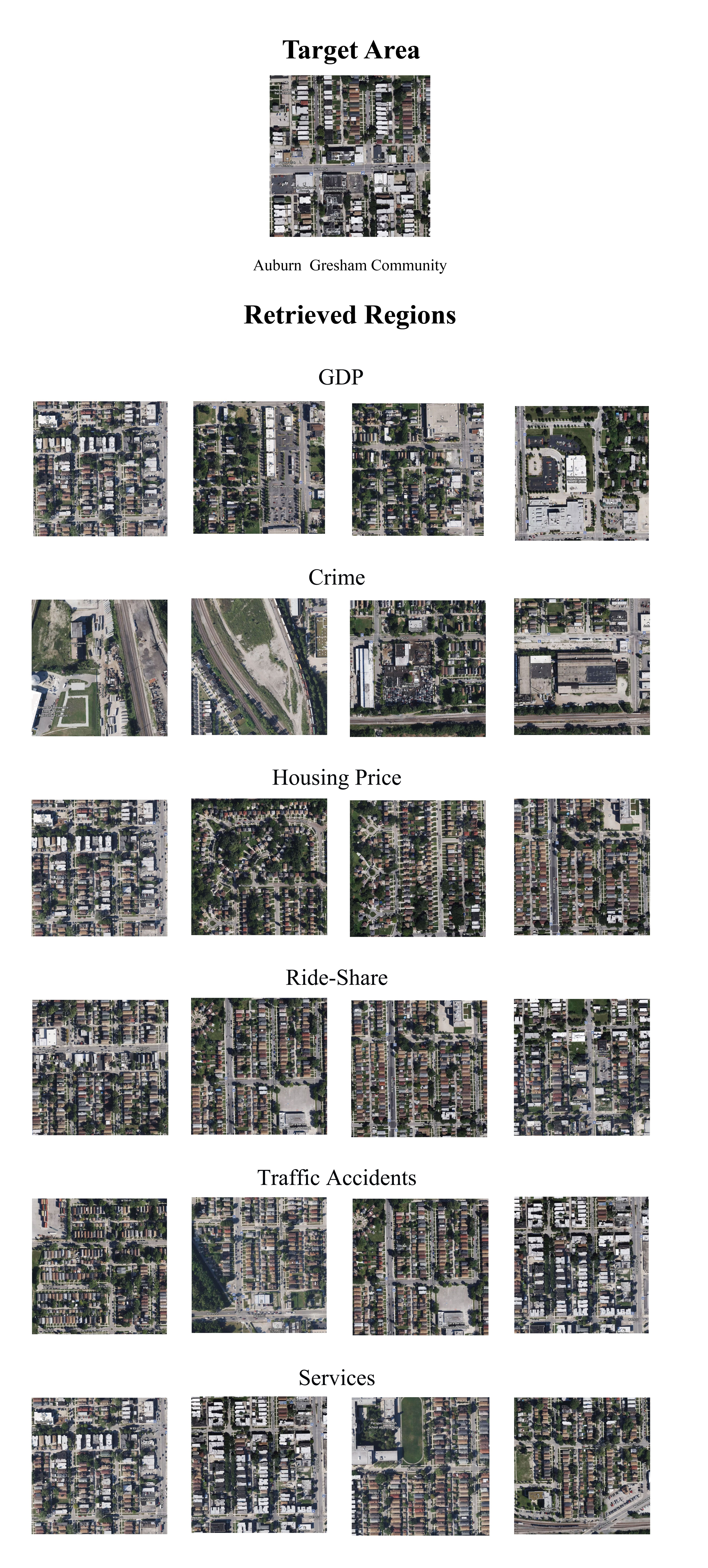}
\caption{An example of dependency retrieval results.}
\label{fig:retrieval_vis}
\end{figure}

\mbox{}\\
To investigate the functionality of the dependency retrieval module, we visualize several top retrieval results for the target region. 
Figure \ref{fig:retrieval_vis} presents an example of a satellite image of the target region in the northwest area of the Auburn Gresham community and four top relevant regions in each task. 

For GDP, the selected areas typically feature organized commercial and institutional land uses, with visible parking lots, banks, supermarkets, and consistent infrastructure layouts. 
Crime-related tiles exhibit fragmented land use patterns, the presence of abandoned warehouses, rail corridors, and poorly maintained lots, all of which contribute to perceived environmental disorder. 
Housing price samples are characterized by well-maintained single-family housing, regular street grids, and abundant greenery, often with curved or cul-de-sac street patterns indicative of higher residential quality.
Ride share hotspots are concentrated in neighborhoods with dense residential fabric intersected by churches, schools, or small commercial outlets, reflecting active human mobility and pick-up/drop-off dynamics.
Traffic accident locations tend to occur near wide arterial roads, intersections, and areas with heavy bus transit activity, reflecting complex traffic flows. 
Finally, service-related tiles show a strong presence of public facilities, such as schools, churches, clinics, and retail plazas, indicating multifunctional neighborhood cores. 
These results showcase our retrieval module's ability to capture urban structures.

\subsubsection{Information Entropy}
\begin{figure}[h]
\centering
\includegraphics[width=0.4\textwidth]{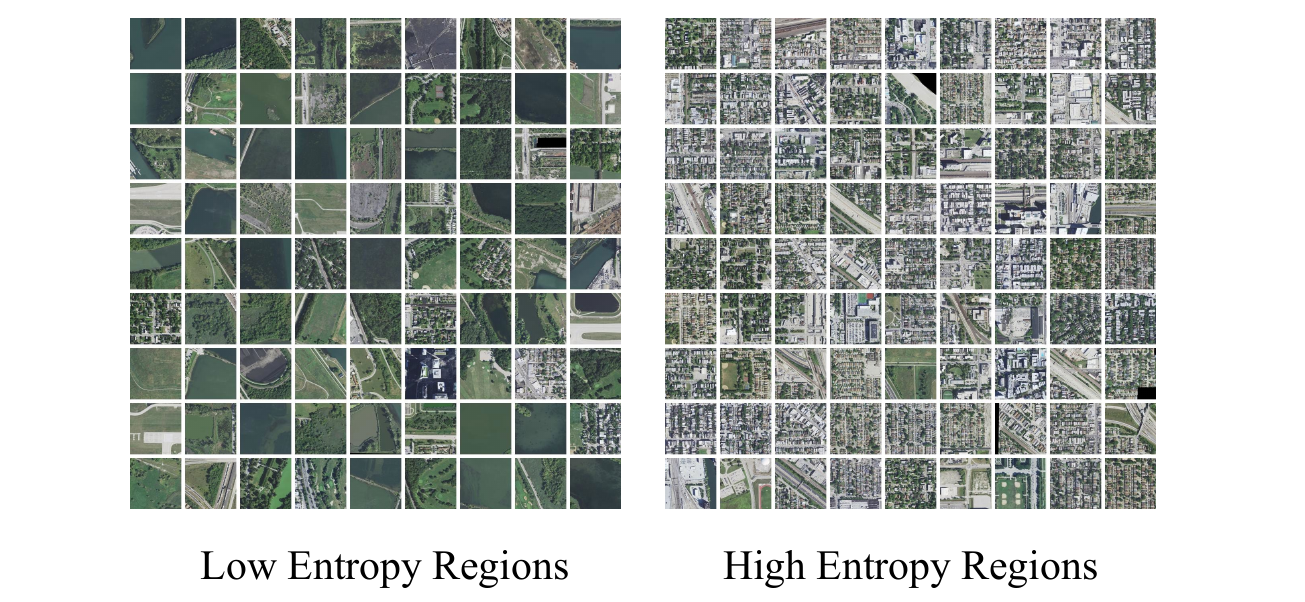}
\caption{A comparison of high and low information entropy regions.}
\vspace{-3mm}
\label{fig:entropy_vis}
\end{figure}

\mbox{}\\
In Section 3.4, we argue that higher information entropy in region representations reflects more complex and diverse urban environments. To support this claim, Figure~\ref{fig:entropy_vis} presents a visual comparison of regions with high and low entropy. As shown, high-entropy regions typically correspond to functionally diverse and detail-rich urban areas, whereas low-entropy regions are often large, homogeneous green spaces. This observation provides qualitative evidence for the effectiveness of using information entropy as a prior in our geospatial attention mechanism.

\subsection{Ablation Study}
To investigate the necessity and effectiveness of each design in our framework, We conduct two sets of ablation studies to evaluate the key contributions of our framework: the task-aware dependency retrieval mechanism and the geospatial attention weighting in GeoTransformer. All experiments are conducted on three representative prediction tasks.

\subsubsection{Dependency Retrieval Mechanism.}  
\mbox{}\\
We compare our task-aware retrieval module against three alternative mechanisms under the same encoder-decoder configuration (DHM + GeoTransformer).  
(1) \emph{Random retrieval} selects context regions uniformly at random from the k-nearest spatial neighbors of the target region.  
(2) \emph{Similarity-based retrieval} uses cosine similarity in the latent feature space to directly retrieve the most similar regions to the target.  
(3) \emph{Sparse retrieval} applies Lasso regression over latent features to identify relevant regions in a task-agnostic manner.
Our full model uses a task-aware prompt passed through a language model to infer the prototype of informative regions. In Table \ref{tab:ablation_retrieval}, results show that our approach consistently outperforms others. For instance, in ride-share demand prediction, task-aware retrieval achieves an average test $R^2$ gain of 0.09–0.20 over other methods. This highlights the value of integrating semantic and task-specific signals in dependency construction.

\begin{table}[h]
\centering
\caption{Ablation results of retrieval mechanisms (R$^2$). All models use DHM encoder and GeoTransformer decoder.}
\begin{tabular}{>{\small}lccc}
\toprule
\textbf{Retrieval Method} & \textbf{GDP} & \textbf{Ride-share} & \textbf{Crimes} \\
\midrule
Random Retrieval & 0.66/0.63 & 0.72/0.70 & 0.60/0.45 \\
Similarity-based Retrieval & 0.75/0.69 & 0.88/0.79 & 0.63/0.51 \\
Sparse Retrieval & 0.79/0.72 & 0.84/0.81 & 0.70/0.46 \\
Task-aware Retrieval (Ours) & 0.81/0.78 & 0.92/0.90 & 0.67/0.56 \\
\bottomrule
\end{tabular}
\label{tab:ablation_retrieval}
\end{table}

\subsubsection{Geospatial Attention Weighting.}  
\mbox{}\\
To isolate the impact of our geospatial weighting design, we conduct an ablation by removing each of the two priors in turn from the GeoTransformer.
(1) \emph{No spatial proximity} removes the $W_S$ term, ignoring distance-based decay.  
(2) \emph{No information entropy} removes the $W_E$ term, treating all retrieved region embeddings as equally informative.  
(3) \emph{No weighting} eliminates both terms, reducing attention to standard cross-attention.  
In Table~\ref{tab:ablation_weighting}, all variants show noticeable performance drops compared to the full model. Among them, removing all weighting yields the largest drop, followed by removing spatial proximity, while removing entropy weighting has the least impact. This confirms the effectiveness of both spatial priors and information richness weighting in guiding attention aggregation.

\begin{table}[h]
\centering
\caption{Ablation results of geospatial attention weighting (R$^2$). All models use DHM encoder and task-aware retrieval.}
\begin{tabular}{>{\small}lccc}
\toprule
\textbf{Weighting Variant} & \textbf{GDP} & \textbf{Ride-share} & \textbf{Crimes} \\
\midrule
No Spatial Weight ($W_S$ off) & 0.77/0.75 & 0.90/0.84 & 0.60/0.59 \\
No Entropy Weight ($W_E$ off) & 0.80/0.75 & 0.88/0.87 & 0.63/0.56 \\
No Weighting & 0.70/0.66 & 0.82/0.73 & 0.59/0.52 \\
Full GeoTransformer (Ours) & 0.81/0.78 & 0.92/0.90 & 0.67/0.60 \\
\bottomrule
\end{tabular}
\label{tab:ablation_weighting}
\end{table}

\section{Conclusion and Discussion}

This study set out to demonstrate that high-dimensional urban forecasting can be streamlined through a unified, modular pipeline that 1) encodes each region once, 2) retrieves task-specific context with a language model, and 3) aggregates everything in an attention-based decoder that can be improved by incorporating spatial priors and informational richness. The experiments confirm that each of those three stages is essential to realizing the goals articulated in the introduction: bridging the gap between graph-based and region-based paradigms, avoiding hand-crafted spatial structures, and remaining plug-and-play for future components.

With a stable encoder in place, the language-driven dependency retrieval module realizes the promise of task specificity that fixed spatial graphs cannot match. For spatially diffuse targets such as GDP and ride-share demand, it consistently surfaces semantically relevant but geographically distant regions and thereby pushes performance well beyond proximity-based or sparsity-based baselines, validating the intuition that LLM reasoning can serve as an automated proxy for expert judgment. GeoTransformer then aggregates target and context information through cross-attention, whose scores are modulated by both physical distance and information entropy. Ablations reported in the study show that removing either prior degrades accuracy, and removing both reduces the decoder to a vanilla GAT with no clear advantage, confirming that the attention design directly addresses the aggregation pitfalls identified in the literature review.

Several limitations open clear avenues for future research. First, our experiments focus on static snapshots; extending the framework to streaming spatiotemporal data will require retrieval and attention mechanisms that evolve in real time. Second, the current prompts are handcrafted; coupling them with domain-specific LLMs or reinforcement-learned prompt generators could deepen the system’s contextual awareness while reducing human effort. Third, although the entropy prior attenuates noisy embeddings, a more explicit uncertainty-aware learning objective may yield further robustness when representations are mined from heterogeneous or low-quality imagery. 

Looking forward, we envisage this work as a stepping-stone toward autonomous geospatial reasoning engines that sit at the heart of urban digital twins. By aligning multimodal sensing, large-scale representation learning, and natural-language reasoning, the framework can evolve into an interactive “what-if” platform: planners describe a policy scenario in natural language, the system retrieves semantically relevant urban context, and GeoTransformer projects multi-domain impacts in one forward pass. With open-source release of code, prompts, and pretrained models, we hope to catalyse a community effort that pushes urban AI from bespoke models toward reusable, modular infrastructure—ultimately equipping cities of varying data maturity to make evidence-based, equitable, and sustainable decisions.

\bibliographystyle{ACM-Reference-Format}
\bibliography{sample-base}
\end{document}